\documentclass[conference, final]{IEEEtran}
\usepackage[noadjust]{cite}

\ifCLASSINFOpdf
  \usepackage[pdftex]{graphicx}
  \graphicspath{{figures/}}
  \DeclareGraphicsExtensions{.png,.pdf}
\else
\fi
\usepackage[cmex10]{amsmath}
\usepackage{amssymb}

\usepackage{array}

\usepackage[tight,footnotesize]{subfigure}
\usepackage{chngcntr}
\counterwithin{paragraph}{subsection}

\usepackage{changes}

\begin{document}
\title{Autonomous Grounding of Visual Field Experience through Sensorimotor Prediction}

\author{
\IEEEauthorblockN{Alban Laflaqui\`ere}
\IEEEauthorblockA{AI Lab, Aldebaran\\
43 rue du Colonel Avia, 75015 Paris, France\\
Email: alaflaquiere@aldebaran.com}
}


\maketitle

\normalem

\begin{abstract}
In a developmental framework, autonomous robots need to explore the world and learn how to interact with it. Without an a priori model of the system, this opens the challenging problem of having robots master their interface with the world: how to perceive their environment using their sensors, and how to act in it using their motors. The sensorimotor approach of perception claims that a naive agent can learn to master this interface by capturing regularities in the way its actions transform its sensory inputs. In this paper, we apply such an approach to the discovery and mastery of the visual field associated with a visual sensor. A computational model is formalized and applied to a simulated system to illustrate the approach.
\end{abstract}


%
\IEEEpeerreviewmaketitle

\section{Introduction}
\label{sec:Introduction}
As advocated by the developmental approach of robotics, a truly autonomous agent should explore its environment and learn on its own how to interact with it \cite{cangelosi2015developmental}. For a \textit{tabula rasa} agent this implies discovering that such an environment does exist, that it has a structure and properties, but also that there exists an interface \cite{hoffman2015interface} to this external world: the body which mediates perception and action.
Having this knowledge emerge in a naive agent represents an intimidating challenge. This is in part why traditional robotic approaches prefer relying on engineers to hand-code this knowledge into robots in the form of sensory processing, decision making, and control algorithms \cite{brady2012robotics}. However, such an a priori definition of perception and action presents limitations and risks. It is too rigid to allow the robot to adapt to unforeseen conditions or to develop new ways to interact with the environment. Moreover it is too complex for the system's designers to encompass every aspect of the rich robot-environment interaction. And above all, engineers' intuitions about the way animals perceive and act might simply be incorrect.

In order to overcome those limitations, one has to tackle the initial challenge of understanding how a naive agent can learn to interact with the world. The problem is the following: how to make sense and use of the uninterpreted sensorimotor flow a naive agent has access to? Without a priori knowledge the incoming sensory flow is like static snow on a screen - furthermore not necessarily encoding an image - and the outgoing motor flow is an unknown choice of actions whose consequences in the world are hidden.
Confronted with this difficulty, the large majority of unsupervised approaches propose to re-encode sensory information based on its statistical properties. This is for instance the case of some currently very successful Deep Learning methods. However, as underlined by the final supervision stage of deep convolutional object recognition \cite{krizhevsky2012imagenet}, those filtered sensory inputs remain unintepreted for the naive agent. It is like the static snow image has been filtered to generate a new (hopefully smaller) one. Noteworthy enough, most of those approaches focus on processing the sensory flow and leave aside the other facet of the problem, the motor flow.

Another approach is required to make initially unintepreted sensory input useful and let a naive agent know how it can act in the world. The sensorimotor approach of perception inspired by the sensorimotor contingencies theory (SMCT) is one promising proposal \cite{o2001sensorimotor}. It claims that perception is \textit{mastering the way sensory inputs can be actively transformed by actions}. Although the original SMCT does span a larger scope of arguments related to cognition and consciousness \cite{oregan2011red}, our sensorimotor approach of perception focuses on the pragmatical aspect of this claim. Perception can be acquired by letting a robot explore the world and discover how its motor commands transform its sensory inputs. Intuitively, this means that a naive agent can explore its motor space and check how it changes the incoming static snow image. The structure in the external world will thus induce structure in the generated transformations. Contrarily to passive sensory processing though, this structure will later be useful for the agent as it describes how it can navigate in its sensorimotor space and allow it to select actions to reach a (sensorimotor) goal. This core idea also blurs the boundary between perception and action, as the agent learns to master its interface with the world, including both how it perceives (how sensory inputs can be actively transformed) and how it acts (the dual: how its actions transform sensory inputs).

Numerous experimental results suggest that such a sensorimotor account of perception is relevant 
\cite{bach2003sensory,kaspar2014experience,witzel2015determines}. In particular, recent results show that humans learn to master their visual field, that is the way ocular saccades transform visual sensory inputs \cite{herwig2014predicting}. More precisely they learn the relations between different sensory inputs encoding the same visual feature at different positions on the retina, as well as the motor commands between them. Such a relation is for instance how a vertical edge is precisely encoded in the fovea but coarsely encoded when projected in the periphery of the retina after a given saccade. It has been shown in artificial setups that modified relations can be learned, leading to an altered visual field experience even after perceptive capacities have stabilized in adult subjects (for instance, associating two periodic visual features of different frequencies in the fovea and the periphery) \cite{herwig2014predicting}.

Taking inspiration from this work and the sensorimotor approach of perception, this paper proposes a formalization of the visual field mastery problem. It is directly in line with previous work in which a simpler setup was considered \cite{laflaquiere2015fov}. Below, a computational model is proposed to describe how sensorimotor transformations related to a moving visual sensor can be discovered and captured by a naive agent. A simulation is introduced to illustrate and evaluate the approach on a visual search task. Finally results are analyzed and current limitations as well as future extensions of the approach are discussed.


\section{Problem formulation}
\label{sec:Problem formulation}

\begin{figure}[!t]
\centering
\includegraphics[width=0.9\linewidth]{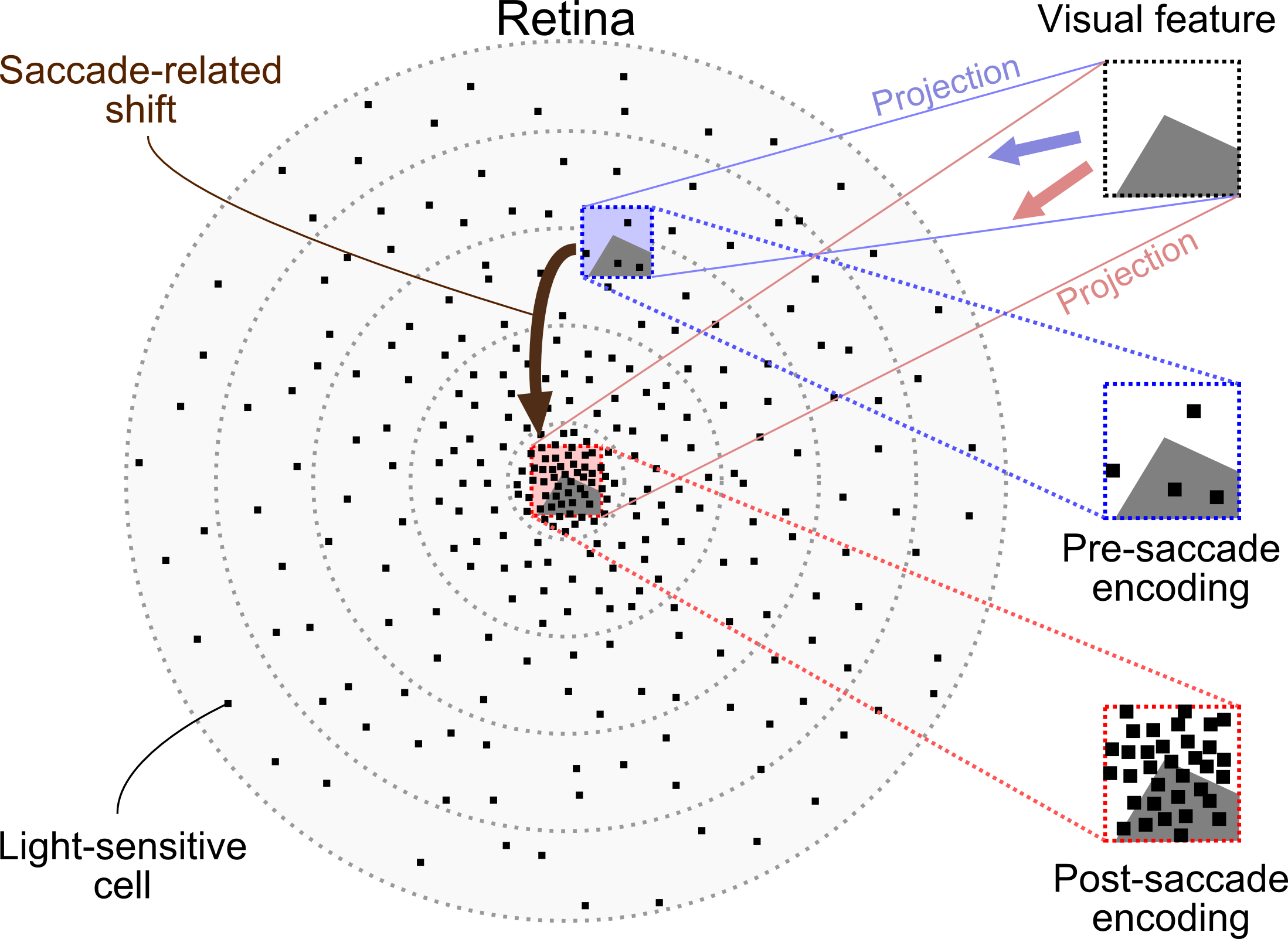}
\caption{Simplified illustration of a retina representing the non-uniform distribution of light-sensitive cells and the resulting variable sensory encoding of visual features. Visual features are projected on the retina and shift when the eye saccades. The sensory encoding of visual features depends on where their projection lands. Resolution is for instance significantly higher in the fovea (center of the retina) than in periphery.}
\label{fig:fig1}
\end{figure}

We are interested in agents equipped with a visual sensor, that is a 2D array of pixels collecting information from a limited part of the environment. Taking inspiration from the human retina, we consider that the sensor array is divided into multiple \emph{receptive fields} gathering a small neighborhood of pixels \cite{lindeberg2013computational}. Each receptive field is considered as an independent sensor generating its own sensory inputs. Note that, like in the human retina, all receptive fields don't necessarily share the same properties: number, relative positions, and nature of light-sensible cells or pixels in our case. The same \emph{visual feature} - visual information coming from a small part of the environment - can thus be encoded differently depending on which receptive field processes it (see Fig.~\ref{fig:fig1}).

Due to the physical embedding of the visual sensor in the world and the structure of the latter, the visual features projected on the retina shift from receptive field to receptive field when the sensor moves (or when the environment moves, which is a dual case we will not consider in this paper). Relations thus exist between the sensory inputs of the different receptive fields and the saccadic motor outputs.
Initially the agent is naive and doesn't know those relations. However, it can discover them by exploring its environment, collecting and modeling sensorimotor experiences shaped by those physical constraints. Doing so, the agent learns to control the interface with the world that the sensor constitutes: it knows which different sensory inputs in different parts of the retina correspond to the same visual feature and how to move visual features in the retina with motor commands.

Formally we denote a receptive field with a superscript $a$ and the sensory state generated by each of them as the multivariate random variable $\mathbf{S}^a$ that can take values:
\begin{equation}
\mathbf{s}^a_i = [s^a_{i,1},\dots,s^a_{i,d^a}],
\end{equation}
where $d^a$ is the number of pixels in receptive field $a$, and $s^a_{i,k}\in\mathbb{R}$ is the individual excitation of the $k^{\text{th}}$ pixel for state $i$.
The same way, we denote the saccadic motor commands the agent can generate with a multivariate random variable $\mathbf{M}$ that can take values:
\begin{equation}
\mathbf{m}_q = [m_{q,1},\dots,m_{q,d^m}],
\end{equation}
where $d^m$ is the number of motors, and $m_{q,k}\in\mathbb{R}$ is the individual excitation of the $k^{\text{th}}$ motor contributing to the sensor movement $q$.
Note that no superscript $a$ is needed as all receptive fields are moved together during saccades. Moreover motors could potentially be redundant regarding the sensor's displacements in the world, in which case the agent would have to discover its actual working space (see \cite{laflaquiere2015learning} for an example of such a sensorimotor structuring).
The naive agents we consider have no specific policy to explore the world. Consequently the agent's exploration strategy is simply set to random commands, similar to motor babbling observed in babies.

We propose to formalize the agent's learning process as the building of a predictive model:
\begin{equation}
P \big( \mathbf{S}^b(t+1)=\mathbf{s}^b_j \;|\; \mathbf{S}^a(t)=\mathbf{s}^a_i \:,\: \mathbf{M}(t)=\mathbf{m}_q \big),
\end{equation}
describing the probability of transition between a pre-saccadic state $\mathbf{s}^a_i$ in receptive field $a$ and a post-saccadic state $\mathbf{s}^b_j$ in receptive field $b$, given the realization of a saccade $\mathbf{m}_q$.
The agent can estimate this probability by exploring the world and compute statistics on sensorimotor transitions data:
\begin{equation}
\Big( \mathbf{s}_i^a(t),\mathbf{m}_q(t)\rightarrow \mathbf{s}^b_j(t+1) \Big).
\end{equation}
Note that the temporal variable $t$ is dropped in future developments to lighten notations. This probabilistic approach fits nicely in the framework of predictive modeling which describes the brain as a predictive machine and can be applied to the capture of sensorimotor contingencies \cite{seth2014predictive}.

\begin{figure*}[!t]
\centering
\includegraphics[width=\linewidth]{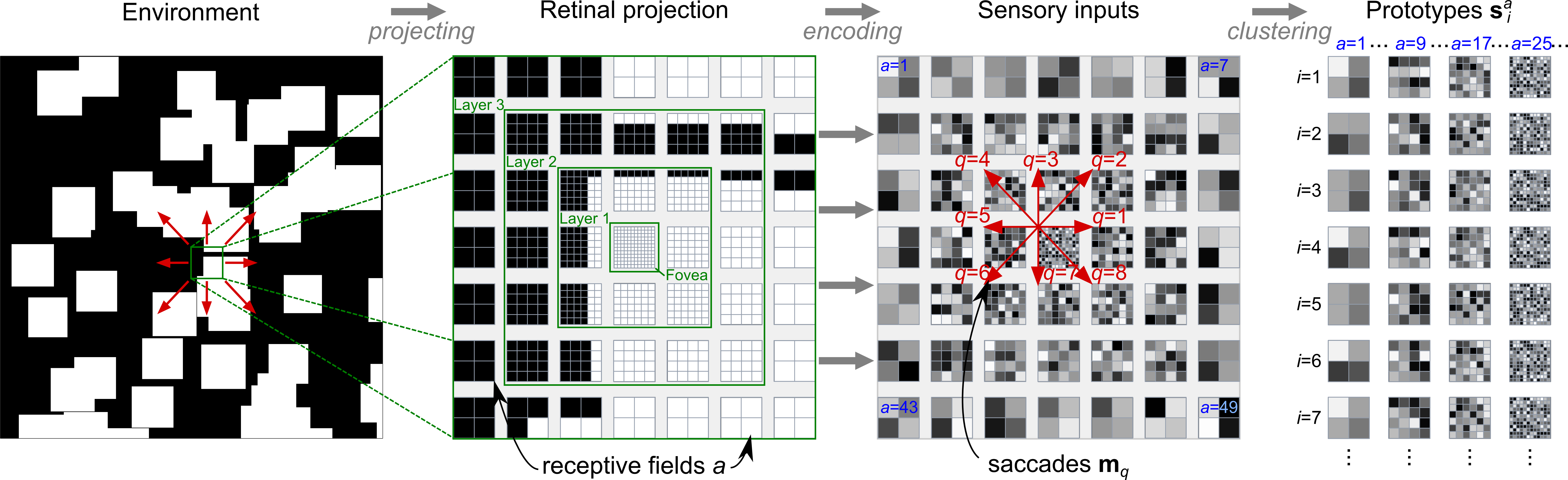}
\caption{The simulated agent is a retina that can move in a simple environment made of white squares on a black background. The retina is divided into $7^2$ receptive fields, organized in 4 layers of artificially decreasing resolutions (from $12^2$ to $2^2$ pixels). The agent can perform saccades that shift the whole visual scene of one receptive field width in 8 directions. The retinal projections are transformed independently in each receptive field to generate sensory inputs. Prototypes are estimated to represent sensory experience in each receptive field using K-means algorithm. }
\label{fig:fig2}
\end{figure*}

The physical embedding of the sensor in the world induce constraints on the agent's experience; these constraints should appear as a structure in the predictive model.
More precisely, specific sensorimotor transitions should be significantly more probable than others as they correspond to visual features shifting between receptive fields during saccades.
To evaluate which receptive fields $(a,b)$ are coupled this way, we propose to evaluate the normalized mutual information between them, given a motor command (random variables are omitted when possible to shorten equations):
\begin{equation}
I(\mathbf{S}^a;\mathbf{S}^b \:|\: \mathbf{m}_q) = \frac{H(\mathbf{S}^b\:|\:\mathbf{m}_q) - H(\mathbf{S}^b\:|\:\mathbf{S}^a,\mathbf{m}_q)}{H(\mathbf{S}^b\:|\:\mathbf{m}_q)},
\label{eq:mutualinformation}
\end{equation}
with $H(\mathbf{S}^b\:|\:\mathbf{m}_q)$ the entropy of receptive field $b$ given the motor command $q$:
\begin{equation*}
H(\mathbf{S}^b\:|\:\mathbf{m}_q) = -\sum_j P\big(\mathbf{s}^b_j\:|\:\mathbf{m}_q\big) \log\Big(P\big(\mathbf{s}^b_j\:|\:\mathbf{m}_q\big)\Big),
\end{equation*}
and $H(\mathbf{S}^b\:|\:\mathbf{S}^a,\mathbf{m}_q)$ its entropy conditioned on the state of receptive field $a$:
\begin{equation*}
H(\mathbf{S}^b\:|\:\mathbf{S}^a,\mathbf{m}_q) = - \sum_{i,j} P\big(\mathbf{s}^b_j,\mathbf{s}^a_i|\mathbf{m}_q) \log\left(\frac{P\big(\mathbf{s}^b_j,\mathbf{s}^a_i|\mathbf{m}_q\big)}{P(\mathbf{s}^a_i|\mathbf{m}_q)} \right).
\end{equation*}
Entropy is a measure of the unpredictability of the post-saccadic variable $\mathbf{S}^b$, that can be conditioned on the outcome of the pre-saccadic variable $\mathbf{S}^a$. Mutual information $I(\mathbf{S}^a;\mathbf{S}^b\:|\:\mathbf{m}_q)$ is thus a measure of how much information $\mathbf{S}^a$ provides about $\mathbf{S}^b$, given a saccade $\mathbf{m}_q$.
For each saccade, mutual information should thus be significantly lower for pairs of receptive fields $(a,b)$ between which visual features shift. Those relations are the way the sensor's physical structure is accessible to the naive agent.

\section{Simulation}
\label{sec:Simulation}

\begin{figure*}[!t]
\centering
\includegraphics[width=\linewidth]{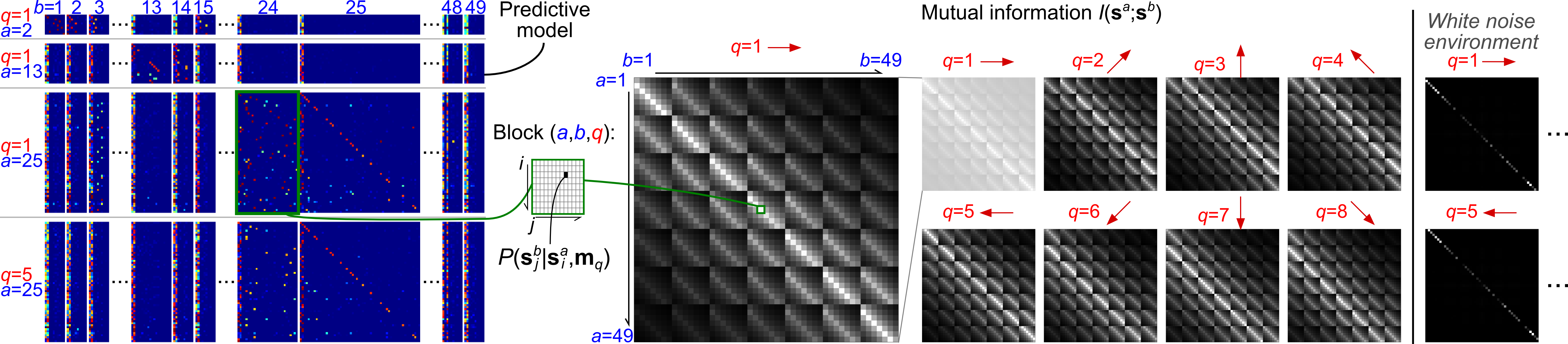}
\caption{[Better seen in color] On the left: Partial presentation of the predictive model estimated by the agent. Each block corresponds to a pair of receptive fields $(a,b)$ and a motor command $q$. Each line $i$ in a block corresponds to the conditional distribution $P\big( \mathbf{S}^b\:|\:\mathbf{s}^a_i,\mathbf{m}_q \big)$ for the given saccade $q$.
On the right: Mutual information $I(\mathbf{S}^a;\mathbf{S}^b\:|\:\mathbf{m}_q)$ computed for each block $(a,b,q)$ and represented as a matrix for each motor command $q$. For each saccade, pairs of receptive fields display high mutual information (white). They are receptive fields between which visual features shift during the corresponding movement. Residual mutual information in other blocks is due to environmental structure and disappear when the agent explores white noise visual scenes (far right).}
\label{fig:fig3}
\end{figure*}

\subsection{System description}
\label{sec:System description}
A simple agent-environment system is simulated in order to apply and evaluate the approach. As illustrated in Fig.~\ref{fig:fig2}, it coarsely captures the kind of interaction a moving eye has with its environment.
The agent is a camera with a $84^2$ pixels retina. This retina is divided into $7^2$ juxtaposed receptive fields of size $12^2$ pixels. Yet, in order to mimic the heterogeneous human retina, the resolution of each receptive field is artificially reduced as it lies further from the center of the retina. This is done by grouping the receptive fields into $4$ concentric layers with respective resolutions of $\big((\frac{1}{6},\frac{1}{3},\frac{1}{2},1)\times 12 \big)^2$ pixels. Practically the resolution is reduced by respectively keeping only one pixel every $(6,3,2,1)$ pixels, both for rows and columns, in the small $12^2$ image received by the receptive fields. This variable resolution reproduces the significant loss of information between the center of our retina - the \textit{fovea} - and its periphery.

Each pixel can originally generate an excitation $s^a_{i,k}$ in the interval $[0,255]$, with the two extremas respectively corresponding to no excitation (black pixel) and full excitation (white pixel). However, in order to emphasize the fact that the agent will not rely on implicit structure in the data, we apply two transformations to disrupt the sensory signal encoding (see Fig.~\ref{fig:fig2} center).
First, different linear transformations $g$ are independently applied to all pixels of the retina to modify their excitation functions:
\begin{equation}
g(x)=\alpha x + \beta, \text{with } \alpha\in[1-\frac{\beta}{255},-\frac{\beta}{255}], \beta\in[0,255]
\end{equation}
The functions' parameters $\alpha$ and $\beta$ are randomly generated for each pixel, and ensure that the transformed input still lies in a subspace of $[0,255]$. Once drawn, the transformation is fixed and considered a property of the sensor (more precisely, of each pixel in the array). Second, the pixels are mixed up before forming the sensory state vector $\mathbf{s}^a_i$. This way the order in which excitations appear in the vector doesn't correspond to any topological organization of pixels in the receptive field. The mixed order is drawn randomly for each receptive field and then considered a fixed property of the sensor (more precisely, of each receptive field).

The agent's sensory experience is discretized by considering that each receptive field $a$ can be in a finite number of prototype states $\mathbf{s}^a_i$. The number $N^a$ of states in receptive fields is arbitrarily set accordingly to their resolutions: $N^a = (\frac{1}{6},\frac{1}{3},\frac{1}{2},1) \times 60$. The states themselves are generated in a data driven way by randomly collecting $2.5\times10^4$ sensory inputs and applying a simple K-means algorithm to cluster them in $N^a$ states $\mathbf{s}^a_i$ for each receptive field (see Fig.~\ref{fig:fig2}).

Likewise a finite number of $Q=8$ saccadic movements $m_q$ are considered. They correspond to the sensor's translations in the retina plane such that the central \emph{foveal} receptive field exchanges positions with the $8$ receptive fields of layer 1 (see Fig.~\ref{fig:fig2}). They have been chosen so that visual features completely shift between receptive fields during saccades. A larger set of such saccades could be considered, for instance by having the fovea exchange position with all others receptive fields. However $Q$ is purposefully kept low to reduce the computational cost of the simulation which would require a parallel porting on GPU to increase efficiency.

Simple artificial environments are generated for the agent to explore. Like illustrated in Fig.~\ref{fig:fig2}, they are images made of a black background on top of which white squares of variable sizes are randomly distributed. Ten different environments are generated this way. During the simulation, the agent successively explores them during the same amount of time.
This environmental simplicity allows us to easily keep track of what is being captured in the predictive model (see Sec.~\ref{sec:Results}). However it shouldn't be considered a drawback of the approach as the focus of this paper lies in capturing the sensor structure, not environmental properties.

\subsection{Estimating the predictive model}
\label{sec:Estimating the predictive model}
In order to estimate the predictive model $P(\mathbf{S}^b\:|\:\mathbf{S}^a,\mathbf{M})$, the agent explores each environment by randomly generating $10^5$ saccades. The number\footnote{Number of saccades $\times$ Number of environments $\times$ Number of pre-saccadic receptive fields $\times$ Number of post-saccadic receptive fields} of individual sensorimotor transitions $(\mathbf{s}^a_i,\mathbf{m}_q\rightarrow\mathbf{s}^b_j)$ experienced by the agent is thus equal to $10^5\times10\times7^2\times7^2 = 2401\times10^6$. The probability of each sensorimotor transition $P(\mathbf{s}^b_j\:|\:\mathbf{s}^a_i,\mathbf{m}_q)$ is then estimated based on those data:
\begin{equation}
P(\mathbf{s}^b_j|\mathbf{s}^a_i,\mathbf{m}_q) = \frac{\text{count}\big( (\mathbf{s}^a_i,\mathbf{m}_q\rightarrow\mathbf{s}^b_j) \big)}{\sum_{j=1}^{N^b} \text{count}\big( (\mathbf{s}^a_i,\mathbf{m}_q\rightarrow \mathbf{s}^b_j) \big)},
\label{eq:estimateP}
\end{equation}
For convenience, those elementary predictive models are stored ``by blocks" in the agent's memory. This way, a block $(a,b,q)$ gathers predictive models related to the receptive fields $a$ and $b$, and the saccade $q$. It forms a small matrix where each row corresponds to a pre-saccadic state $i$ and each column to a post-saccadic state $j$ (see Fig.~\ref{fig:fig3}). According to Eq.~\eqref{eq:estimateP}, each row of the matrix thus defines a conditional distribution $P\big( \mathbf{S}^b\:|\:\mathbf{s}^a_i,\mathbf{m}_q \big)$ over the post-saccadic states $\mathbf{s}^b_j$.

\begin{figure*}[!t]
\includegraphics[width=0.93\linewidth]{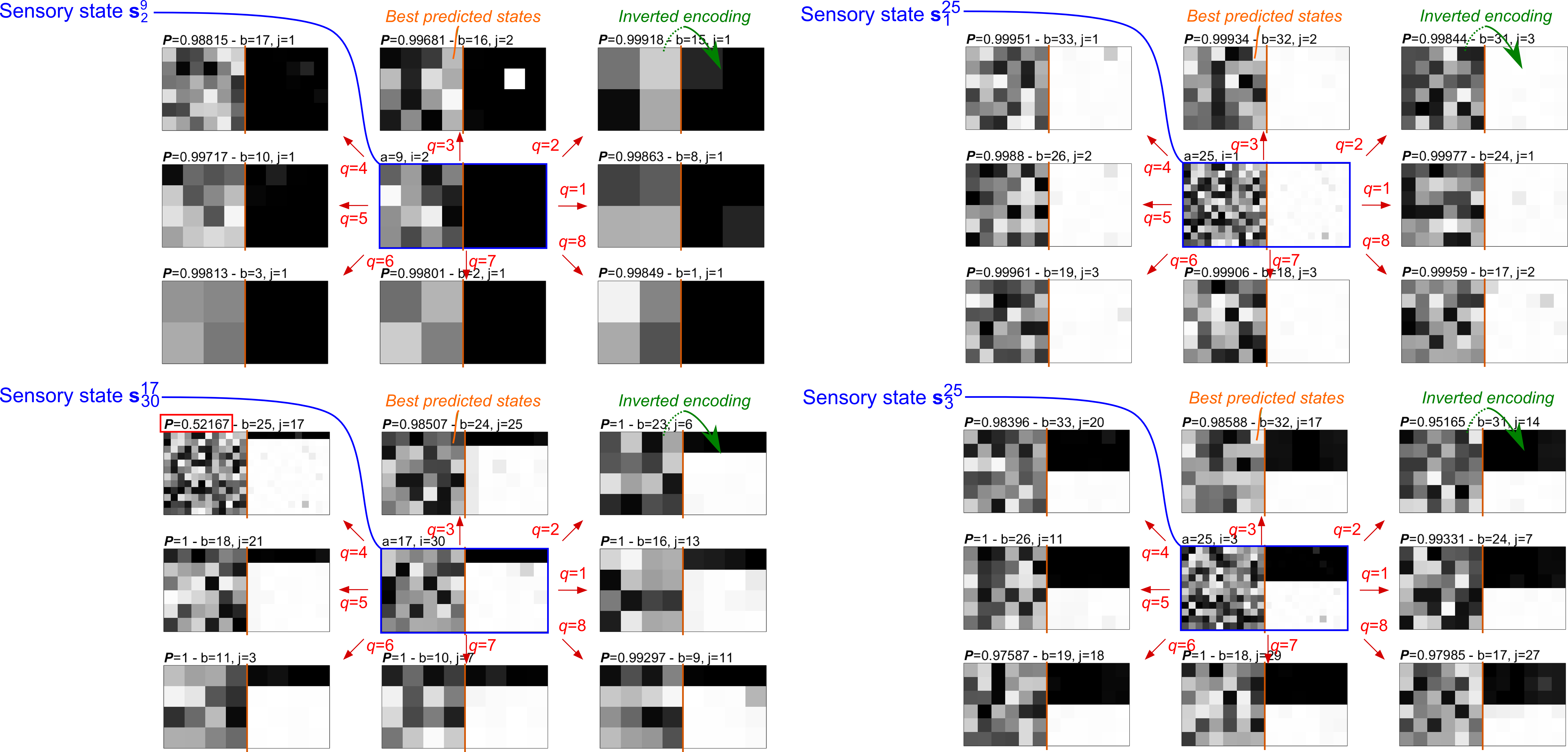}
\caption{Four pre-saccadic sensory states $\mathbf{s}^a_i$ surrounded by the most probable
states $\mathbf{s}^b_j$ they get transformed into by each saccade $q$. The latter states are determined by considering the receptive field $b$ with the highest mutual information $I(\mathbf{S}^a;\mathbf{S}^b|\mathbf{m}_q)$ for a given saccade $q$ and selecting the $j$ with the highest probability $P(\mathbf{s}^b_j\:|\:\mathbf{s}^a_i,\mathbf{m}_q)$ (displayed on top of the patch). For the sake of visualization, the visual feature it encoded by each sensory state is estimated, by inverting the sensory encoding of the corresponding receptive field, and displayed on its right.}
\label{fig:fig4}
\end{figure*}

\subsection{Visual search}
\label{sec:Visual search}
A visual search task is proposed in order to illustrate how the estimated sensorimotor predictive model is useful to the agent. The latter is placed in an environment similar to the ones explored during the learning process. Repeatedly, a desired foveal sensory input $\widehat{\mathbf{s}}^b_j$ (equivalent to a visual feature) is defined. The agent has to counterfactually search for it in the field of view and perform a saccade so that the desired sensory state is reached.
Practically, the desired sensory inputs $\widehat{\mathbf{s}}^b_j$ are selected by looking at the visual features (un-encoded ground truth) received by other receptive fields and encoding them as if they were projected in the fovea. This ensures that the desired visual feature is present in the current field of view and that the desired sensory state can potentially be reached. Moreover, because only saccades of one receptive field width have been considered during the learning, only the layer 1 of receptive fields directly surrounding the fovea is considered during this search (see Fig.~\ref{fig:fig2}).

The agent estimates which motor command to perform as follows. First for each peripheral receptive field $a$ a motor command $\mathbf{m}^a_{f}$ is determined such that:
\begin{equation}
f = \text{argmax}_q \big( I(\mathbf{S}^a;\mathbf{S}^f|\mathbf{m}_q) \big),
\end{equation}
where $\mathbf{S}^f$ corresponds to the post-saccadic foveal sensory state. The motor command $\mathbf{m}^a_{f}$ is thus the one that maximizes the mutual information between the receptive field $a$ and the fovea. From an external point of view, $\mathbf{m}^a_{f}$ is the saccade that makes visual features shift from the receptive field $a$ to the fovea.
Second the saccade $\mathbf{m}^{a^*}_{f}$ the agent has to perform to foveate the desired visual feature is determined as:
\begin{equation}
\mathbf{m}^{a^*}_{f} \;|\; a^* = \text{argmax}_a P\big(\widehat{\mathbf{s}}^b_j\:|\:\mathbf{s}^a_i,\mathbf{m}^a_{f}\big).
\end{equation}
Intuitively the agent selects the receptive field $a$ whose current sensory state $\mathbf{s}^a_i$ has the highest probability to transform into the desired foveal state $\widehat{\mathbf{s}}^b_j$ after performing the saccade $\mathbf{m}^{a^*}_f$.

\section{Results}
\label{sec:Results}

\subsection{Predictive model and sensor structure}
As claimed in Sec.~\ref{sec:Problem formulation}, the physical embedding of the sensor in the world should be translated in the predictive model as a structure of highly predictable sensorimotor transitions. As illustrated in Fig.~\ref{fig:fig3}, this structure can indeed be observed in the block matrices $(a,b,q)$ of the predictive model: some blocks (for instance $(a=25,b=24,q=1)$) display stronger predictability patterns than others (for instance $(a=25,b=49,q=5)$). This can be formally measured and more easily visualized by looking at the normalized mutual information computed for each block $(a,b,q)$ according to Eq.~\eqref{eq:mutualinformation}. Depending on the executed saccade $\mathbf{m}_q$ different pairs of receptive fields $(a,b)$ display a significantly higher mutual information than others (see Fig.~\ref{fig:fig3}). From our omniscient point of view, we can confirm that those pairs correspond to receptive fields between which visual features shift during the corresponding saccade. The mutual information matrix displays regular patterns for each saccade because we purposefully organized its rows and columns according to the receptive fields order in the retina. Note however that the agent doesn't have access to such knowledge and cannot take advantage of those patterns.

Yet we can also observe that mutual information between receptive fields is not binary - a spectrum of intermediary values does exist. This variability is due to the fact that the predictive model not only captures structure induced by the visual sensor but also structure induced by the environment. Because the environment statistically displays local continuity, neighboring receptive fields can better predict the state of a target receptive field. Intuitively receiving a uniform black visual feature in a receptive field allows for instance the agent to fairly accurately predict that neighboring receptive fields also receive a uniform black feature.
Such a continuity property could be taken advantage of to estimate a topological map of the different receptive fields, as was proposed in \cite{kuipers2008drinking}. However, although it can help an external observer visualize the data, we see no incentive for a naive agent to build such a map.
To demonstrate the impact of the environmental structure on the predictive model, the simulation has been run with a random noise environment (each pixel is initially drawn in $[0,255]$). Figure~\ref{fig:fig3} shows how the lack of environmental structure removes intermediary values of mutual information to only leave the ones related to the actual sensor structure.

In the coupled blocks $(a,b,q)$ induced by the sensor structure, the estimated predictive model also informs the agent on which sensory states $\mathbf{s}^a_i$ correspond to the same visual feature encoded in different parts of the retina. Figure~\ref{fig:fig4} shows a few example sensory states $\mathbf{s}^a_i$ and the different sensory states $\mathbf{s}^b_j$ they predict in the coupled receptive fields $b$ given the $8$ motor commands. Along the sensory inputs the agent has access to are displayed the actual visual features they correspond to (computed by inverting the sensory transformation described in Section.~\ref{sec:System description}). One can observe that each group of those sensory states encodes the same visual feature, which can nonetheless be encoded with different resolutions. The agent can thus estimate that completely different sensory inputs actually correspond to the same visual feature (information about the world). Moreover, he knows which motor action transforms one into the other, which from an external point of view corresponds to making the visual feature shift on the retina.
Inverting the sensory encoding also reveals why two sensory inputs are predicted most of the time in uncoupled blocks $(a,b,q)$ (see for instance block $(a=13,b=25,q=1)$ in Fig.~\ref{fig:fig3}). They correspond to the uniform white and black features which are significantly more probable in these artificial environments. When receptive field $a$ doesn't inform about receptive field $b$, predicting those two features is a safe estimate.

Finally, one can also notice that, because of the resolution loss between the different retina layers, the association between lower resolution pre-saccadic states $\mathbf{s}^a_i$ and higher resolution post-saccadic ones $\mathbf{s}^b_j$ can be ambiguous. Intuitively, it simply corresponds to the fact that a blurry pattern can correspond to multiple sharp ones. This can be seen in the second panel of Fig.~\ref{fig:fig4} where sensory state $\mathbf{s}^{17}_{30}$ transforms into the foveal state $\mathbf{s}^{25}_{17}$ after saccade $\mathbf{m}_4$: the probability is only $\approx \! 0.52$ because another visual feature (white two black rows at the top) can also be predicted with a probability of $\approx \! 0.48$.
Note that the opposite is not true as higher resolution patterns can unambiguously predict their lower resolution counterpart. We argue that it is the rationale why an initially naive agent would naturally prefer the foveal encoding of a visual feature compared to every other one in the retina: it is the only encoding that can unambiguously predict all other ones. That is also the reason why foveation seems like a sensible objective in the visual search task.

The visual search task described in Sec.~\ref{sec:Visual search} was performed on $10^3$ successive iterations. The agent succeeded in reaching the desired foveal state in every trial.
This remarkable success rate is due to the way the visual search task is designed: the desired sensory state is foveal (unambiguous) and can be reached at each iteration. However it highlights the quality of the estimated predictive model which always associates paired sensory states in different receptive fields. This simple search task illustrates how successfully the agent can use the predictive model it estimated to control its initially unstructured sensorimotor interaction with the environment.

\section{Discussion}
\label{sec:Discussion}
This paper proposed a sensorimotor formalization of the problem of having a naive agent autonomously learn to master a visual field. A computational model has been defined and applied on a simulated system to illustrate how an agent can discover the regular transformations induced on its sensorimotor experiences by the physical embedding of its sensor in the world. Those transformations define both how different sensory inputs coming from different parts of the sensor encode the same visual features, and how motor commands transform some into the others. Those regularities are discovered by exploring the world and can later be used by the agent to internally simulate interactions with the world and consequently select the most favorable action to perform, like illustrated in a simple visual search task.
The simulation also showed how specific sensory states can naturally emerge among the different ones that encode the same visual feature in an heterogeneous sensor. In a retina-like sensor,  this is the case of foveal encodings that can accurately predict all others peripheral encoding due to their higher resolution.

Future extensions of the approach should illustrate other core aspects of the sensorimotor approach of perception. In particular, we'll address the problem of autonomous emergence of semantics. So far, the simulated agent identified different sensory states that encode the same visual features, which can be seen as a first step towards semantics. However, a more convincing result will be to show how different features can be autonomously grouped together based on more abstract properties (for instance group together all vertical or horizontal edges in the proposed simulation). Those properties can indeed be identified in the way the corresponding sensory states can be transformed through action.
Future developments will also illustrate how sensorimotor transformations can be organized in a hierarchical model that can be latter used for an efficient exploration of new environments.
Finally, the algorithm proposed in this paper will be ported on a GPU to benefit from parallel computing in order to evaluate a more realistic retina exploring more complex environments.

\bibliographystyle{IEEEtran}
\bibliography{bibICDL}

\end{document}